\documentclass{article}
\usepackage{ijcai24}

\usepackage{times}
\usepackage{soul}
\usepackage{url}
\usepackage[hidelinks]{hyperref}
\usepackage[utf8]{inputenc}
\usepackage[small]{caption}
\usepackage{graphicx}
\usepackage{amsmath}
\usepackage{amsthm}
\usepackage{booktabs}
\usepackage{algorithm}
\usepackage[switch]{lineno}
\usepackage[table,dvipsnames]{xcolor}

\title{A Generative Approach to Surrogate-based Black-box Attacks}

\author{Raha Moraffah \and Huan Liu \\
\affiliations
         Arizona State University
         \emails{\{rmoraffa, huanliu\}@asu.edu}}
\usepackage{hyperref}

\usepackage{amsmath,amssymb,amsfonts}
\usepackage{graphicx}
\usepackage{textcomp}
\usepackage{xcolor}

\usepackage{multirow}
\usepackage{booktabs}

\usepackage{textcomp}
\usepackage{xcolor}

\newcommand{\m}{GSBA}
\DeclareMathOperator*{\argmin}{\arg\!\min}

\usepackage{colortbl}
\definecolor{Gray}{gray}{0.85}
\newcolumntype{a}{>{\columncolor{Gray}}c}
\usepackage{diagbox}

\usepackage{algorithm}
\usepackage{algpseudocode}
\usepackage{graphicx}
\usepackage{textcomp}
\usepackage{xcolor}

\usepackage{subfigure}
\usepackage{caption}
\begin{document}

\maketitle

\begin{abstract}
Surrogate-based black-box attacks have exposed the heightened vulnerability of DNNs.  These attacks are designed to craft adversarial examples for any samples with black-box target feedback for only a given set of samples. State-of-the-art surrogate-based attacks involve training a \textit{discriminative} surrogate that mimics the target's outputs. The goal is to learn the decision boundaries of the target. The surrogate is then attacked by \textit{white-box} attacks to craft adversarial examples similar to the original samples but belong to other classes.  With limited samples, the discriminative surrogate fails to accurately learn the target's decision boundaries, and these surrogate-based attacks suffer from low success rates.
Different from the discriminative approach, we propose a \textit{generative} surrogate that learns the distribution of samples residing on or close to the target's decision boundaries. The distribution learned by the generative surrogate can be used to craft adversarial examples that have imperceptible differences from the original samples but belong to other classes. 
The proposed generative approach results in attacks with remarkably high attack success rates on various targets and datasets. %Particularly, our attack achieves +15\% improvement over existing state-of-the-art attacks.

%simple yet powerful algorithm that replaces traditional discriminative surrogates with the generator component (a generative surrogate).

%State-of-the-art surrogate-based attacks consist of two components: a generator and a surrogate that is trained on the data sampled from the generator to imitate the black-box target. The surrogate is then attacked by white-box attacks to craft adversarial examples. To effectively generate the data, the generator learns the distribution of samples residing extremely close to or on the target's decision boundaries, which includes adversarial examples. Intuitively, %compared to the surrogates trained on only a portion of the data sampled from the generator, 
%the generator itself contains more information than the surrogate about the potential adversarial examples. Therefore, we hypothesize that we can directly exploit the generator's distribution to generate adversarial examples. To validate the hypothesis, we propose a simple attack using only the generator to directly craft adversarial examples from the generator's distribution by drawing the closest examples from the other classes to the benign examples of the original class. Since the adversarial examples reside extremely close to or on the target's decision boundaries, the drawn samples from the generator's distribution are highly likely to be adversarial.
%Our novel approach results in attacks with remarkably high attack success rates on various targets and datasets. Particularly, our attack achieves +15\% of improvement over existing state-of-the-arts attacks, in just one attempt.
\end{abstract}

\section{Introduction}
\label{sec:intro}
% Adversarial attacks 
% blackbox 
% surrogate-based
% Existing solutions and peoblems

The widespread deployment of Deep Neural Networks (DNNs) in various real-world applications~\cite{chitta2021neat,kasar2016face} has raised concerns about their reliability. One major concern is DNN's vulnerability to adversarial examples that are crafted by adding human-imperceptible perturbations to original samples and are misclassified by the DNNs~\cite{szegedy2013intriguing}. Designing strong attacks in practical settings allows assessing the extent of DNN vulnerability.
%Designing attacks is one way of improving defense. 
White-box attacks that assume full knowledge about the target DNN such as its architectures and weights have been developed~\cite{goodfellow2014explaining,carlini2017towards}. Since this setting is impractical, black-box attacks are being studied, which can only access the target through queries.

Black-box attacks are divided into two main types, namely, query-based and surrogate-based attacks. Query-based attacks make multiple queries to the target for each sample to be attacked to estimate its adversarial directions during the attack~\cite{ilyas2018prior,moraffah2022query}. Despite their high success rates, these attacks have limited real-world practicality due to requiring queries proportional to the number of attacked samples, which is not feasible in a real-world setting due to limited inference time or monetary limits~\cite{ilyas2018prior,ma2021simulating}.

\begin{figure}
    \centering
    \includegraphics[width=20pc, height = 10pc]{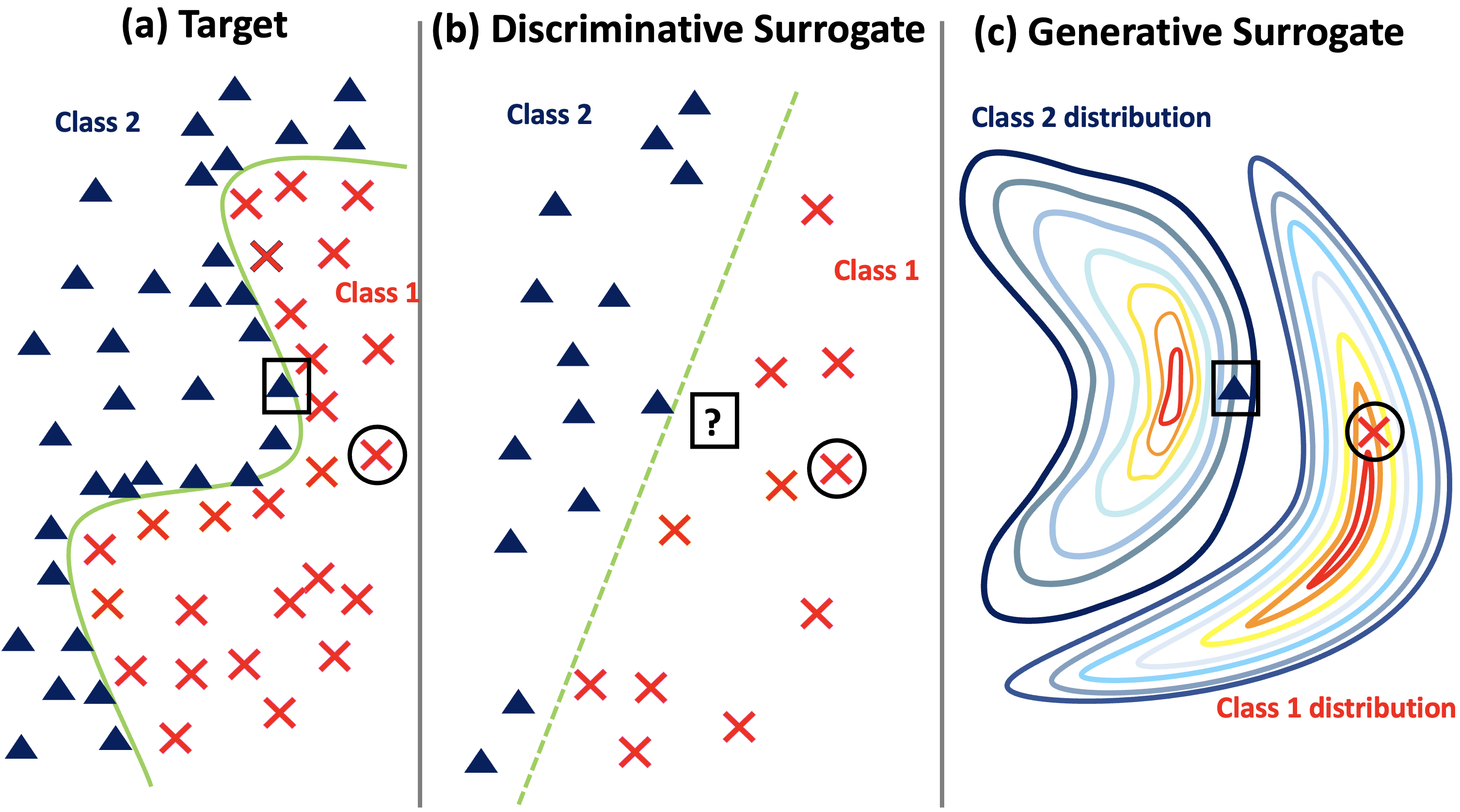}
    \caption{Illustration of the attack on discriminative vs generative surrogates. Figure~(a) shows the target decision boundary for two classes. Figure~(b) illustrates the discriminative surrogate that aims to learn the target decision boundary with limited samples. Figure~(c) demonstrates a generative surrogate that learns the distribution of samples residing on/close to the target decision boundary. The adversarial example (in $\square$) for the original sample (in $\circ$) cannot be identified using the discriminative surrogate, due to its wrong decision boundary, 
    whereas in the generative surrogate, the adversarial example is directly sampled from the distributions.}
    \label{fig:motivating}
\end{figure}

Surrogate-based attacks, on the other hand, provide a more practical attacking scenario. These attacks are designed to attack any samples with only access to the target feedback for a given set of samples. Existing surrogate-based attacks train \textit{discriminative} surrogates to imitate the target outputs. The goal is to train surrogates that learn the target's decision boundaries, which are exploited by white-box attacks to craft adversarial examples. Failing to accurately learn the target's decision boundaries with limited samples, these attacks obtain low success rates.

In this paper, we propose a novel surrogate-based attack that trains a \textit{generative} surrogate, which learns the distribution of samples on/close to the target's decision boundaries. 
This distribution should contain adversarial examples that are extremely close to their original samples but belong to a different class than the original class. We then propose a novel attack to utilize this generative surrogate and explore its distribution to identify adversarial examples. %HL: Our proposed attack identifies the adversarial examples such that they are close to the original examples of the original class (i.e., have imperceptible differences) but belong to the distribution of samples on any class but the original class decision boundary (untargeted attack) or the distribution of samples on the target class's decision boundaries (targeted attack). 
As shown in Figure~\ref{fig:motivating}, the discriminative surrogate fails to learn the target's decision boundaries and find the adversarial examples; whereas an attack on the generative surrogate does not rely on the decision boundaries but utilizes the generative surrogate's distribution to identify adversarial examples.

%In particular, our proposed attack identifies samples generated by the surrogate, which are close to the original examples of the original class (i.e., have imperceptible differences) but are assigned to different classes (in the case of the untargeted attack) or belong to a pre-specified target class (in the case of the targeted attack). Figure~\ref{fig:motivating} demonstrates the difference between the discriminative and generative surrogates and how the attacks are conducted using each one of them.
Extensive experiments show that our novel attack significantly outperforms state-of-the-art surrogate-based attacks under the same setup (attack norm) with efficiency. %HL: In addition to its effectiveness, our proposed attack is also extremely efficient, as it generated adversarial examples in just one step, as opposed to existing attacks that use multiple iterations. 
Our contributions are summarized as follows:
\begin{itemize}
    \item We propose a generative surrogate that models the distribution of potential adversarial examples to replace the existing discriminative surrogates, 
    \item  We design an effective and efficient attack with the generative surrogate to craft adversarial examples, and 
    \item We conduct extensive experiments for comparative performance evaluation. % results demonstrate that our attack achieves remarkably higher success rates in just one step, even in a more difficult targeted setting. 
    
\end{itemize}

%Particularly, as illustrated in Figure~\ref{fig:motivating}, under the untargeted setting, that only requires the adversarial example to be misclassified, we use the most similar example generated by the surrogate for any other class except the original as adversarial examples.  Under the targeted setting, we return the closest example from the target class as the adversarial example.

\section{Related Work}
%\subsection{Adversarial Attacks}
Several adversarial attacks have been developed under the white-box setting, where the adversary has full knowledge of the target and utilizes its gradients to generate the adversarial examples~\cite{szegedy2013intriguing,goodfellow2014explaining,madry2017towards,carlini2017towards}. Restricted by the strong assumption of full access to the target, this setting is far from realistic. To craft more practical adversarial examples, 
black-box adversarial attacks, which access the target only via queries, have been proposed. These attacks are classified into query-based and surrogate-based attacks. Query-based attacks estimate the gradients of the target on the fly and for each sample during the attack~\cite{ma2021simulating,du2019query,li2019nattack,chen2017zoo}. These attacks require querying the target numerous times per sample, making them inappropriate for real-world settings. Surrogate-based attacks train a surrogate network for the black-box target to mimic its behavior and attack the surrogate to generate adversarial examples. 
These attacks only require querying the target during the surrogate training and are the only type of attacks applicable to the case where no queries are allowed during the attack. 
Existing efforts on surrogate-based attacks have been focused on learning more representative surrogates to improve the success rate of these attacks by selecting/learning more useful samples for querying the target~\cite{papernot2017practical,orekondy2019knockoff,zhou2020dast,wang2021delving,sun2022exploring}. However, all of these methods train a discriminative surrogate that mimics the target outputs for the given set of samples. Our proposed framework, on the other hand, trains a generative surrogate to learn the distribution of potential adversarial examples and proposes a single-step attack to craft adversarial examples using this generative surrogate. Recently, Moraffah et al. propose to probe the target's joint distribution of inputs and outputs to generate adversarial examples~\cite{moraffah2022exploring}. Though both adopt a generative surrogate, their purposes differ:  the existing work learns the entire target distribution for every pair of input and output, which requires an exponential number of queries, and our generator is restricted to learning the distribution of samples on/close to decision boundaries for adversarial attacks.

\section{Methodology}
% \subsection {Threat Model}
% In this section, we discuss the threat model and outline the adversary’s background knowledge and the goal of the attack.

%HL\subsection{Attack Scenario}
%Black-box attack
%probability and label only scenario
% talk about access to the dataset
% no query is allowed during the inference
Our attack is conducted in a black-box setting in which the attacker has no knowledge of the target model (e.g. target's architecture, parameters, weights, etc.), and can only access the target via making queries and observing the target's response. For the target response, there are two scenarios: (1)  Probability-only Scenario: the target returns the output class probabilities, denoted with  ``-P'' suffix; and (2) Label-only Scenario: The target returns the output labels of the classes for the queries samples, denoted with ``-L'' suffix. Since our attack is framed in a surrogate-based black-box setting, the attacker is only allowed to make a limited number of queries to the target to attack all samples.

\subsection{Problem Statement} \label{prelim}

Let $\mathcal{C}(.) \colon x \in [0, 1]^d \to y \in \mathbb{R}^c$ be a target classifier, and $(x, y)$ denote a pair of input image and its true label. An adversarial attack perturbs $x$ into $x_{adv}$, its corresponding adversarial example, such that it has an imperceptible difference from $x$ and is misclassified by the $\mathcal{C}$. Formally:

\begin{equation}
\begin{aligned}
\label{eq:advobj}
& ~~\argmin_{x_{adv}} f(\mathcal{C}(x_{adv}), t)\\
%&\text{s.t.} ~~ f(x_{adv})\geq 0,%, x_{adv}\in [0, 1]^d,
& \textbf{s.t. } \|x_{adv} - x\|_p \leq \delta,
\end{aligned}
\end{equation}

\noindent where $f$ is the adversarial objective that measures the degree of uncertainty of $\mathcal{C}$ in assigning $x_{adv}$ to class $t$. Common choices of $f$ are classification loss~\cite{goodfellow2014explaining} and C\&W loss function~\cite{carlini2017towards}. 
Under the untargeted setting, the classification loss is defined as $-\log p_{\mathcal{C}}(t\mid x_{adv})$., where $t=y$ is the true label of the image. In the C\&W loss, the $f$ is measured by $\max\{0, \log \mathcal{C}(x_{adv})_{y} - \underset{i \neq y}{\max} \log(\mathcal{C}(x_{adv})_i )\}$, where $\mathcal{C}(x_{adv})_i$ is the i-th probability output of the target. 
Under the targeted setting, the classification loss is defined as $\log p_{\mathcal{C}}(t\mid x_{adv})$, $t$ is the predefined target class, and $f(x_{adv}, t) = \max\{0,  \underset{i \neq t}{\max} \log(\mathcal{C}(x_{adv})_i - \log \mathcal{C}(x_{adv})_{t})\}$ is the C\&W loss. $l_p$ norm ($\| . \|_p$) is used to measure the difference between the original and adversarial examples. Under the surrogate-based black-box setting, the target is unknown is only accessible through querying the target and getting (input, output) pairs for a limited number of samples and not every sample to be attacked. 

% theoritical analysis
%\subsection{Overview}
\subsection {Proposed Framework}
We explain our proposed \textbf{G}enerative \textbf{S}urrogate-based \textbf{B}lack-box \textbf{A}ttack ({\m}) that utilizes the distribution of potential adversarial examples, learned by a generative surrogate, to craft adversarial examples. {\m} consists of two steps: (S1) Generative Surrogate Training - to train a generative surrogate that learns the distribution of potential adversarial examples; and (S2) Attack Strategy: to directly utilize the surrogate's distribution and generate adversarial examples. % in one step.  
% In the following, we elaborate on the details of each step.
 
\subsubsection{S1: Generative Surrogate Training} \label{generativesurrogatetraining}
\begin{figure*}
\centering
%\hspace{-0.5cm}
\includegraphics[width=\textwidth, height = 15pc]{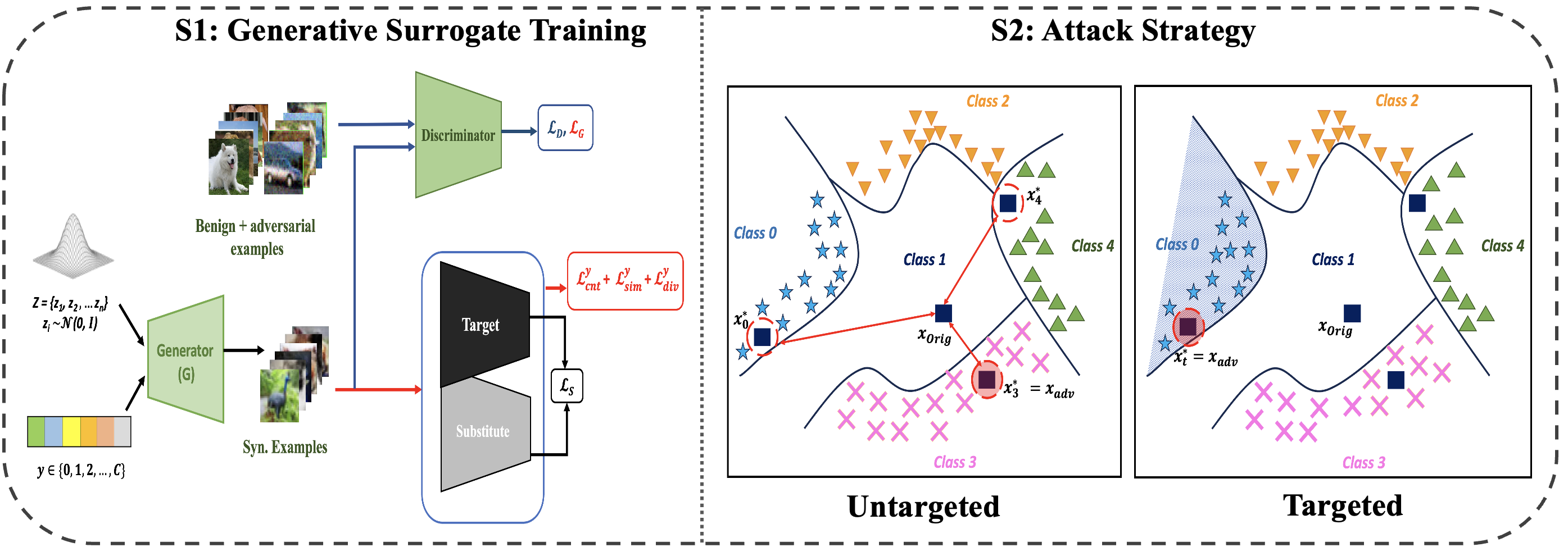}
\caption{An overview of the {\m} framework. (1) \textbf{S1: Generative Surrogate Training.} This step aims to train a surrogate that learns the distribution of samples with three properties: (i) realness ($\mathcal{L}_{\operatorname{G}}$, $\mathcal{L}^{adv}_{\operatorname{G}}$, and $\mathcal{L}^{y}_{\operatorname{cnt}}$);(ii) high inter-class similarity ($\mathcal{L}^{y}_{\operatorname{sim}}$); and (iii) high intra-class diversity ($\mathcal{L}^{y}_{\operatorname{sim}}$). Once trained, the surrogate will be directly leveraged to generate adversarial examples.; (2) \textbf{S2: Attack Strategy.} The original benign sample is shown with $\blacksquare$. Under the untargeted setting, first the most similar samples to the original sample generated by the surrogate for any other class except the original adversarial examples are identified (denoted by $x^*_i$). The sample with minimum distance from the original sample is then selected as the adversarial example ($x^*_{3}$).  Under the targeted setting, the most similar sample from the target class (shown in hatch) is selected as the adversarial example ($x^*_{0}$). }
 \label{gentrain}
%\hspace{-0.5cm}
\end{figure*}

This step trains a generative surrogate that learns the distribution of potentially adversarial examples, i.e., samples that reside extremely close/on the target's decision boundaries. Such examples possess three characteristics: (1) they are realistic-looking: adversarial examples should have imperceptible difference from the original real samples by definition; (2) they have high inter-class similarity: adversarial examples reside extremely close to/on the decision boundaries of the target model; and (3) they have high intra-class diversity: the distributions of potential adversarial examples should be capable of generating samples that cover all intersections of decision boundaries for all classes. This is important for the targeted attack setting which requires the adversarial examples to be misclassified as a specific target class, i.e., cross the decision boundary between the original and targeted class. The generator training step is depicted in Figure~\ref{gentrain}. Next, we explain different losses designed to enforce each property on the generative surrogate.

\textbf{Learn the distribution of realistic samples}
 % GAN loss
 % label controablle
 % adversarial training
To do so, %learn the distribution of realistic-looking samples, 
we adopt a generative architecture based on Generative Adversarial Network (GAN)~\cite{goodfellow2014generative}. To generate samples from given classes, our architecture takes a random noise $z$ and a class label $y \in\{0,1,2 \dots C\}$ ($C$ is the number of classes) and outputs $X = G(z, y)$. To make the examples realistic, the following adversarial objectives are optimized: 
\begin{equation}
\begin{aligned}
\label{eq:gan}
& \mathcal{L}_{\operatorname{D}} = \underset{{x} \sim
 \mathbb{P}_d} {\mathbb{E}}[\log \operatorname{D}(x)] + \underset{{x} \sim
 \mathbb{P}_g} {\mathbb{E}}[\log (1 - \operatorname{D}(\operatorname{G}(z, y))], \\
 & \mathcal{L}_{\operatorname{G}} = \underset{{x} \sim
 \mathbb{P}_g} {\mathbb{E}}[\log (1 - \operatorname{D}(\operatorname{G}(z, y))]. \\
\end{aligned}
\end{equation}

 To ensure the class label of the generated samples, we extend our loss with a class-controlling loss:
 \begin{equation}
\label{eq:control}
\mathcal{L}^y_{cnt} = \operatorname{CE}(\operatorname{T(\operatorname{G}(z, y))}, y), 
\end{equation}
 where CE is the cross-entropy loss, T is the target network and $y$ are the labels fed to the generator.

Optimizing for Eqs.~\eqref{eq:gan} and~\eqref{eq:control}, the generator learns the distribution of realistic looking benign examples. Since the generator is supposed to represent the distribution of potential adversarial examples, it should also learn the characteristics of realistic-looking adversarial examples. To this end, we propose to augment the training with adversarial examples. Specifically, given a set of benign examples $X$ we craft their corresponding adversarial examples $X_{adv}$ using the BIM~\cite{kurakin2018adversarial} attack. We then use these examples to train the generator via optimizing $\mathcal{L}^{adv}_{\operatorname{D}}$ and $\mathcal{L}^{adv}_{\operatorname{G}}$ as defined in Eq.~\eqref{eq:gan}.

However, since the target is a black-box, we are unable to use it for class-controlling loss or generating adversarial examples. Instead, we use a substitute network that distills the target's knowledge and mimics its behavior. We adopt a knowledge distillation loss, which can be used to force the substitute to imitate the target's output on a given set of samples~\cite{zhou2020dast,wang2021delving,zhang2022towards}. Formally, given a set of samples $X$ generated by the generator $G$, we minimize the distance between the substitute and target's outputs:

 \begin{equation}
\label{eq:surrogatetraining}
\mathcal{L}_S = \operatorname{dist}(S(X), T(X)),
\end{equation}
where $\operatorname{dist}(.)$ is a function to measure the distance between the Surrogate S and the target T's outputs on samples $X$. In the Label-only scenario, the distance is measured via the Cross Entropy (CE) loss between the class labels produced by the target and surrogate, while in the probability-only scenario, it will be measured via the $l_2$-norm of the difference between the probability outputs $\| T(X) - S(X) \|_2^2$.

\textbf{Inter-class Similarity}
% C and W loss
 To model the distributions of samples on/close to the target's decision boundaries, we maximize the inter-class similarity loss. The inter-class similarity loss measures the distance of the samples from their corresponding decision boundaries. This is equivalent to measuring how close the sample is to being misclassifed as the nearest class across the decision boundary. Since the C\&W loss objective~\cite{carlini2017towards}  measures the degree to which the sample is likely to cross the decision boundary and be misclassified, we adapt the C\&W loss objective to measure the inter-class similarity:

 \begin{equation}
\label{eq:interclass}
\mathcal{L}^y_{sim} = \underset{j \neq y}{\max} \log \operatorname{S}(x)_j - \log \operatorname{S}(x)_{y},
\end{equation}
 where $\operatorname{S}(x)_j$ is the j-th probability output of the S, and $y$  the original class label.
Minimizing Eq.~\eqref{eq:interclass} decreases the sample's probability of being classified as the original class by pushing it toward the nearest class label's decision boundary. 
%\raha{original target is replaced with the surrogate?}

\textbf{Intra-class Diversity}
% Entropy loss
Ideally the distribution of potential adversarial examples for each class is scattered across decision boundaries between that class and all other classes. Optimizing for Eq.~\eqref{eq:gan}, Eq.~\eqref{eq:control}, and Eq.~\eqref{eq:interclass} characterizes the label-controlled distributions of samples near the decision boundaries but does not promote the disperseness of generated samples. An additional loss is thus required to enforce the disperseness of the samples of the distribution. Optimizing the inter-class similarity objective (Eq.~\eqref{eq:interclass}), pushes the samples close to the decision boundary to have nearly-equal highest and second-highest probabilities of being assigned to a class, while the rest of probabilities are close to zero. If samples of each class are evenly distributed across its decision boundary, it means on average equal number of samples reside close to each intersection of the original class's boundary and other classes. In other words, the second-highest probability assigned by the model to the samples (probability of the other class that shares the decision boundary) on average are equally distributed across all class categories except for the original class. To mathematically measure this, 
 we propose to maximize the information-entropy of a vector of average probabilities of all classes except for the original one:% a widely-used measure of diversity of probability distributions~\cite{}, of all other classes except for the original class  to force the generated samples to evenly distribute across all classes as follows:

%The information entropy of the probability vector $P=\{p_1, p_2,\dots p_n\}$ defined as $\mathcal{H}(P) = -\frac{1}{n}\sum_i^np_i\log p_i$, has been widely served as a measure of diversity~\cite{}. 

 \begin{equation}
\label{eq:intradiv}
\mathcal{L}^y_{div} = \mathcal{H}(\frac{1}{N}\sum_i^N S^i_{c:0\dots C\neq y}(x)),
\end{equation}
where $\mathcal{H}(P) = -\frac{1}{K}\sum_i^K p_i$ is the information-entropy of a probability vector  $P=\{p_1, p_2, \dots p_k\}$, and  $S^i_{c:0\dots C\neq y}(.)$ is the output probability vector of the surrogate $S$ for sample $i$ without the highest class probability.

\textbf{Generator Optimization}
Combining the Eq.~\eqref{eq:gan}, Eq.~\eqref{eq:control}, Eq.~\eqref{eq:interclass}, and Eq.~\eqref{eq:intradiv}, the final objective of the generator's to learn the class-controlled distribution of samples scattered close to target's decision boundaries is as follows:

\begin{equation}
\label{eq:finalg}
\mathcal{L}_G =(\mathcal{L}_{\operatorname{G}} + \mathcal{L}^{adv}_{\operatorname{G}}) + \alpha_1 \mathcal{L}^y_{cnt} + \alpha_2 \mathcal{L}^y_{sim} + \alpha_2 \mathcal{L}^y_{div},
\end{equation}
where in our experiments $\alpha_1=1$, $\alpha_2=1$, and $\alpha_3=1$. 
%The discriminator will be optimized using the $\mathcal{L}_{\operatorname{D}} 
% + \mathcal{L}^{adv}_{\operatorname{D}} $%the Eq.~\eqref{eq:gan}.

\subsubsection {S2: Attack Strategy}
%talk about problem of implicit distribution
%provide solution--> reconstruction loss of GAN
% final attack objective (plug and play) emphasize 
% talk about the untargeted as well as the targeted attack

In this section, we explain our novel simple attack strategy,  as illustrated in Algorithm~\ref{algo:learning}. Our attack directly exploits the generative surrogate (trained in the previous section) to craft the adversarial examples. The intuition behind our attack is that since the surrogate represents the distribution of examples residing extremely close to/on target's decision boundaries, which includes adversarial examples, it can be directly utilized to generate the adversarial examples. Specifically, depending on the attack's setting (untargeted or targeted) our attack draws the closest sample to the original sample from the generator's distribution that belongs to a different class than the original class or belongs to a pre-specified target class. To learn the closest sample from the generator's distribution, we propose to minimize a reconstruction-based loss that minimizes the distance between the closest sample generated by the generator $G$ and the original example:

 \begin{equation}
\label{eq:closest}
\begin{gathered}
z^* = \argmin_{z^*} \|G(z^*, y) - x_{orig}\|,\\
x^*_y = G(z^*, y)
\end{gathered}
\end{equation}

Depending on the attack setting, the criterion to identify the adversarial example will be different. Next, we formulate our attack under the untargeted and targeted settings.

\textbf{Untargeted Attacks:}
An adversarial example is only required to be misclassified to a class different from the original example's class.  Hence, we identify the closest sample ${x^*_y}$ from all classes $y=0, \dots C $ except for the original class, to the original example (Eq.~\eqref{eq:closest}); the adversarial example is the one with the smallest distance to $x_{orig}$: %select the one with minimum distance to the generated adversarial example   

 \begin{equation}
\label{eq:untargeted}
x_{adv} = \argmin_{y=0, \dots C} \|x - x^*_y \|.
\end{equation}

  \begin{algorithm}[ht]
    \caption{{\m} Attack Strategy}\label{algo:learning}
    \hspace*{\algorithmicindent} \textbf{Input:} Trained surrogate G (from {\m}'s step (1)), original sample $x_{orig}$, Untargeted, perturbation budget $\delta$, target class t, C number of classes.\\
    \hspace*{\algorithmicindent} \textbf{Output:} The adversarial example $x_{adv}$.
    \begin{algorithmic}[1]
    \If{Untargeted == True} \Comment{Untargeted Attack}
                %\State $grad = \mathcal{M}_\theta(x_{adv})$
                \State $x^*_y \leftarrow \argmin \|G(z^*, y) - x_{orig}\|$,  y= 0, 1, \dots C 
                \Comment{Find the closest sample using Eq.~\eqref{eq:closest}}
                %HL \Comment{Find the closest sample generated by G's distribution for all $C$ classes using Eq.~\eqref{eq:closest}}
                \State $x_{adv} \leftarrow \argmin_{y=0, \dots C} \|x - x^*_y \|$
                \Comment{Eq.~\eqref{eq:untargeted}}
                
                 \If{$\|x_{adv} - x_{orig} \| \leq \delta$} 
                 \Comment{Perturbation budget}
                 %HL\Comment{Check for the perturbation budget criterion}
                  \State \Return  $x_{adv}$
                 \Else 
                 \State \Return  None
                 \EndIf

        %\State Sample $w \sim U(-1,1)^c$
    \Else \Comment{Targeted Attack}
        \State ${x^*_t} = \argmin \|G(z^*, t) - x_{orig}\|$ \Comment{Find the closest sample using Eq.~\eqref{eq:closest}}
        %HL \Comment{Find the closest sample generated by G's distribution for class $t$ using Eq.~\eqref{eq:closest}}
        \State $x_{adv} \leftarrow x^*_t$  
        \If{$\|x_{adv} - x_{orig} \| \leq \delta$} \Comment{Perturbation budget}
        %HL\Comment{Check for the perturbation budget criterion}
                  \State \Return  $x_{adv}$
                 \Else 
                 \State \Return  None
                 \EndIf
    \EndIf
    
    % \State\Return  $x_{adv}$
    %\textbf{Output}  \thata of meta model $\mathcal{M}$
    %\RETURN{}Parameters  of meta model $\mathcal{M}$
    \end{algorithmic}
    \end{algorithm}
    
\textbf{Targeted Attacks:}
The adversarial example is required to be misclassified as a pre-specified target class $t$ which is different from the original class. Therefore, the adversarial example will be the nearest example to the original sample but belong to the target class, i.e., $x^*_t$, where $t$ is the adversarial target class and $x^*_t$ is generated via Eq.~\eqref{eq:closest}.

 In the final step for both untargeted and targeted settings, the identified adversarial examples are checked for the perturbation budget criterion ($\|x_{adv} - x_{orig}\|_p \leq \delta$), where $\delta$ is the perturbation budget. The attack will be considered a success only if the criterion is met.

\section{Experiments}
 We examine three main aspects of the {\m}: (1) {\m}'s performance compared with the SOTAs in both untargeted and targeted settings; (2) Ablation study of {\m} in terms of modeling the distributions of potentially adversarial examples (Surrogate Ablation Study) and utilizing the information provided by this distribution to generate adversarial attacks compared to other attacks (Attack Ablation Study); and (3) Performance of the {\m} w.r.t. training query budgets.
In the following, we first explain our experimental setup and then discuss our experimental results. More implementation details and experimental results can be found in the Appendix.
%In addition, we also showcase adversarial examples generated by our attack. In the following, we first explain our experimental setup. We then discuss our experimental results.
\subsection{Experimental Setting}
%In the following, we explain the setup of our experiments.

\textbf{Dataset and Target Models}\label{datatarget}
We utilize two widely-used datasets, namely CIFAR-10~\cite{krizhevsky2009learning} and CIFAR-100~\cite{krizhevsky2009learning}. Following previous research on black-box attacks~\cite{moraffah2022exploring,ma2021simulating}, we randomly select 1,000 \textit{correctly classified} samples from the validation set of each dataset to evaluate the attacks and fix them across different experiments. For the target architectures, we adopt three commonly used models for CIFAR-10, i.e., AlexNet~\cite{krizhevsky2009learning}, ResNet-20~\cite{he2016deep}, and VGG-16~\cite{simonyan2014very}. For CIFAR-100, we adopt ResNet-50~\cite{he2016deep} and VGG-19~\cite{simonyan2014very} as used by the SOTAs~\cite{wang2021delving,zhang2022towards}.

\begin{table*}[ht]
\caption{Experimental results of untargeted attack on CIFAR-10 and CIFAR-100 datasets.} \label{table:mainuntar}
\centering
\small
\resizebox{\textwidth}{!}{\begin{tabular}{c|c|@{\extracolsep{1pt}}*{6}{c}|@{\extracolsep{1pt}}*{4}{c}}
\toprule   
 &\multirow{3}{*}{}  & \multicolumn{6}{c}{CIFAR-10}  & \multicolumn{4}{c}{CIFAR-100}\\

 \cmidrule(r){3-12}

  Type&Attack & \multicolumn{3}{c}{Attack Success Rate ($\uparrow$)} & \multicolumn{3}{c}{AVG. Steps ($\downarrow$)}   & \multicolumn{2}{c}{Attack Success Rate ($\uparrow$)} &\multicolumn{2}{c}{AVG. Steps ($\downarrow$)}\\
 %\cmidrule{2-4}
\cmidrule{3-7}
\cmidrule{8-9}
\cmidrule{10-12}
&&AlexNet& ResNet-20 &  VGG-16 & AlexNet& ResNet-20 &  VGG-16  & Resne-50& VGG-19& Resne-50& VGG-19\\
 \midrule
  \multirow{7}{*}{\rotatebox{90}{Probability-Only}}& {\m-P}&\textbf{91.7\%}& \textbf{89.8\%} &\textbf{ 92.3\%}&  \textbf{ 1} & \textbf{ 1}& \textbf{ 1}  &\textbf{99.7\%} & \textbf{99\%} &    \textbf{ 1} & \textbf{ 1} \\ 
  \cmidrule{2-12}
  &TDB-P~\cite{moraffah2022exploring}&75\% & 83.5\% & 80.1\%&219 &215 & 258& 78.2\% & 74.1\%&248 & 241\\
 \cmidrule{2-2}
 &DFTA-P~\cite{zhang2022towards}&62.8\% & 77.5\%  & 75.6\%  &    228 & 248  & 273 &  70.4\%  & 59.9\% & 305  & 279  \\
 
 &ST-Data-P~\cite{sun2022exploring}&65.2\%&72.8\%&73.3\%
  &241&254&256&64\%&56.2\%&299&286\\
&DAST-P~\cite{zhou2020dast}& 50.4\% & 49\% & 51.9\%  &     289 & 301& 295    & 69.7\% & 50.4\% & 278& 341 \\
\cmidrule{2-2}
 &Knock-off-P \cite{orekondy2019knockoff}&36.9\% & 28.5\% & 29.1\%  &313& 348& 413&26.5\% & 25.0\% & 442 & 475\\
   &JBDA-P~\cite{papernot2017practical}&33.8\%&24.0\%&18.1\%&318&341&462&22.3\%&19.1\%&456&492\\

  \midrule

\multirow{7}{*}{\rotatebox{90}{Label-Only}}& {\m-L}&\textbf{89.2\%} & \textbf{88.8\%}  & \textbf{88.9\%} &   \textbf{1} & \textbf{1}& \textbf{1} &\textbf{98.8\%}  & \textbf{99.2\%} & \textbf{1}& \textbf{1}  \\ 
  \cmidrule{2-12}
  &TDB-L~\cite{moraffah2022exploring}&74.7\% & 80.5\% & 81.2\%& 227 &218 & 258& 76.1\% & 75.1\%&252 & 236\\
  \cmidrule{2-2}
 &DFTA-L~\cite{zhang2022towards}&62.9\% & 68.6\%  & 76.6\% &    249 & 273  & 295  &73.1\%  & 57.2\%   & 266  & 304   \\
 &ST-Data-L~\cite{sun2022exploring}&63.9\%&70.0\%&71.3\%&253&267&281
  &65.8\%&58.2\%& 312&298\\
    
  &DAST-L~\cite{zhou2020dast} &53.3\% & 47.1\%  & 52.2\% &     325 & 249& 282  & 62.2\%  & 47.8\% & 261& 424   \\
  \cmidrule{2-2}
 &Knock-off-L \cite{orekondy2019knockoff}&32.4\% & 23.1\% & 28.7\%  &307& 349& 418&22.1\% & 25.0\% & 442 & 479\\

   &JBDA-L~\cite{papernot2017practical}&33.6\%&22.9\%&18.5\%&316&354&472&21.1\%&18.1\%&455&494  \\

\bottomrule
\end{tabular}}
\end{table*}

\begin{table*}[ht]
\caption{Experimental results of targeted attack on CIFAR-10 and CIFAR-100 datasets.} \label{table:maintar}
\centering
\small
\resizebox{\textwidth}{!}{\begin{tabular}{c|c|@{\extracolsep{1pt}}*{6}{c}|@{\extracolsep{1pt}}*{4}{c}}
\toprule   
 &\multirow{3}{*}{}  & \multicolumn{6}{c}{CIFAR-10}  & \multicolumn{4}{c}{CIFAR-100}\\
 \cmidrule(r){3-12}
  Type&Attack & \multicolumn{3}{c}{Attack Success Rate ($\uparrow$)} & \multicolumn{3}{c}{AVG. Steps ($\downarrow$)}   & \multicolumn{2}{c}{Attack Success Rate ($\uparrow$)} &\multicolumn{2}{c}{AVG. Steps ($\downarrow$)}\\
 %\cmidrule{2-4}
\cmidrule{3-7}
\cmidrule{8-9}
\cmidrule{10-12}
&&AlexNet& ResNet-20 &  VGG-16 & AlexNet& ResNet-20 &  VGG-16  & Resne-50& VGG-19& Resne-50& VGG-19\\
 \midrule
  \multirow{7}{*}{\rotatebox{90}{Probability-Only}}& {\m-P}& \textbf{55.0\%} & \textbf{63.5\%} & \textbf{62.4\%} &   \textbf{1} & \textbf{1} & \textbf{1}& \textbf{59.5\%} & \textbf{58.5\%} &   \textbf{1} & \textbf{1}     \\ 
  \cmidrule{2-12}
 &TDB-P~\cite{moraffah2022exploring}&42.3\%&42.1\%&47.8\%&232 & 298& 329 &39.7\%&38.8\%&300& 421\\
 \cmidrule{2-2}
 &DFTA-P~\cite{zhang2022towards}&30.9\% & 40.8\%  & 38.3\% &  271 & 370 & 355 & 33.3\%  & 32.7\% &   388 & 465  \\
 
 &ST-Data-P~\cite{sun2022exploring}&35.6\%&38.7\%&42.1\%&293 & 355& 371&32.4\%&30.3\%& 366 & 463\\

  &DAST-P~\cite{zhou2020dast}&35\% & 37.2\% & 36.1\% &    354 & 320 & 358 & 30.4\% & 30.1\%&    398 & 481  \\
   \cmidrule{2-2}
 &Knock-off-P \cite{orekondy2019knockoff}&21.4\%&16.3\%&17.6\%&421 & 429 & 453&18.3\%&13.1\%&562 & 637  \\
   &JBDA-P~\cite{papernot2017practical}&16.2\%&14.9\%&14.7\%&440 & 473 & 498 &14.8\%&8.2\% &581 & 681\\
  \midrule
\multirow{7}{*}{\rotatebox{90}{Label-Only}}& {\m-L}& \textbf{55.7\%} & \textbf{64.3\%} & \textbf{54.3\%} &   \textbf{1} & \textbf{1} & \textbf{1} & \textbf{56.5\%}& \textbf{53.5\%} &   \textbf{1} & \textbf{1}  \\ 
  \cmidrule{2-12}
  &TDB-L~\cite{moraffah2022exploring}&40.9\%&41.2\%&45.7\%&   238 & 302& 341     &39.1\%&35.7\%& 303& 428\\
\cmidrule{2-2}

 &DFTA-L~\cite{zhang2022towards} &30.0\% & 35.5\%  & 38.4\%  &  244 & 339 & 349  & 30.2\%  & 30.4\% &   398 & 502   \\
 &ST-Data-L~\cite{sun2022exploring}&35.1\%&34.9\%&39.1\%&294 & 351& 381&32.7\%&28.9\%&364 & 471 \\
 
  &DAST-L~\cite{zhou2020dast} &36.1\% & 39.1\%  & 38.8\% &   241& 368 & 342 &30.3\%  & 30.2\%  &   368 & 490   \\
  \cmidrule{2-2}
 &Knock-off-L \cite{orekondy2019knockoff}&20.8\%&14.9\%&17.3\%&427 & 429 & 469&17.1\%&11.4\%&568 & 637 \\
   &JBDA-L~\cite{papernot2017practical}&14.1\%&13.7\%&14.9\%&442 & 471 & 506&14.8\%&7.6\%&581 & 685 \\
\bottomrule
\end{tabular}}
\end{table*}

\textbf{Evaluation Metric} We report the Attack Success Rate (ASR) and the average number of steps (i.e., attack iterations) taken to generate successful examples (AVG. Steps). The ASR is the ratio of the number of successfully generated adversarial examples to the number of correctly classified samples in the evaluation set. The AVG. Step is served as a measure of the efficiency of the attack, where lower values imply a smaller running time and a higher attack efficiency. 

\begin{table*}[ht]
\caption{Results of Ablation Study on CIFAR-10 and CIFAR-100 datasets on VGG-16 and ResNet-50, respectively. } 
\centering
\small
\resizebox{\textwidth}{!}{\begin{tabular}{c|c|cccc|cccc}%{l@{\extracolsep{1pt}}*{21}{c}}
\toprule  
 % \diagbox{Type}{Data}&&&&&&&&&&&&&&&&&&&&\\
 \diagbox{Type}{Data}&Attack&\multicolumn{4}{c}{CIFAR-10}&\multicolumn{4}{|c}{CIFAR-100}\\
 \midrule
 \multirow{5}{*}{\rotatebox{90}{Probability-Only}}&&$ASR_{untar}$&$AVG. Step_{untar}$&$ASR_{tar}$&$AVG. Step_{tar}$&$ASR_{untar}$&$AVG. Step_{untar}$&$ASR_{tar}$&$AVG. Step_{tar}$\\
 \cmidrule{2-10}
  &Surrogate+PGD&73.9\%&261&50.9\%&345&82.7\%&257&44.2\%&309\\
\cmidrule{2-2}
 &Base&62.2\%&1& 41.1\%&1&76.0\%&1&38.2\%&1\\
 &+$\mathcal{L}_G^{adv}$&67.9\%&1& 43.9\%&1&78.2\%&1&45.5\%&1\\
 &+$\mathcal{L}^y_{sim}$&86.4\%&1& 53.6\%&1&93.1\%&1&54.1\%&1\\
\rowcolor{Gray}&+$\mathcal{L}^y_{div}=${\m}&\textbf{92.3\%}&\textbf{1}&\textbf{62.4\%}&\textbf{1}&\textbf{99.7\%}&\textbf{1}&\textbf{59.5\%}& \textbf{1}\\
\midrule
 \multirow{4}{*}{\rotatebox{90}{Label-Only}}&Surrogate+PGD&75.7\%& 262&51.2\%&351&85.2\%&264&44\%&316\\
\cmidrule{2-2}
&Base&54.6\%&1&36.5\%&1&71.5\%&17&30.6\%&1\\
 &+$\mathcal{L}_G^{adv}$&58.1\%&1&40.9\% &1&78.3\%&1&32.8\%&1\\
 &+$\mathcal{L}^y_{sim}$&82.2\%&1&49.9\% &1&93.1\%&1&43.1\%&1\\
\rowcolor{Gray}&+$\mathcal{L}^y_{div}=${\m}&\textbf{88.9\%}&\textbf{1}&\textbf{54.3\%}&\textbf{1}&\textbf{98.8\%}&\textbf{1}&\textbf{56.5\%}&\textbf{1} \\

\bottomrule
\end{tabular}}
\label{table:ablation}
\end{table*}

\subsubsection{Compared State-of-the-art Methods} \label{compmethod}
%\textbf{State-of-the-arts:}
We adopt three types of state-of-the-arts: (i) \textit{Attacks with generative surrogate:} \textbf{TDB}~\cite{moraffah2022exploring}, the only attack in this category, trains a generative surrogate to mimic the joint distribution of the target on (input, output) pairs; (ii) \textit{Attacks with discriminative surrogates trained on the generator data}: These attacks use the generator to generate the samples to query the target and craft a dataset to train the discriminative surrogate. The surrogate generates the same output as the target for the given dataset. The difference between these methods originates from the quality of the data generated by their generator. \textbf{DFTA}~\cite{zhang2022towards}, \textbf{ST-Data}~\cite{sun2022exploring}, and \textbf{DAST}~\cite{zhou2020dast} are the SOTAs in this category; and (iii) \textit{Attacks with discriminative surrogates trained on traditionally selected data:} these attacks augment or select data from a given dataset to train a discriminative surrogate. \textbf{JBDA}~\cite{papernot2017practical} and \textbf{Knock-off}~\cite{orekondy2019knockoff} are two representative methods in this category. Note that our proposed {\m} differs from all aforementioned baselines since it utilizes a generative surrogate. The surrogate is different from TDB's as it is concentrated on the distribution of samples close to/on the target's decision boundaries and not the entire target distribution.

\subsection{Comparison with State-of-the-art attacks}
We conduct experiments in untargeted and targeted settings.

\textbf{Untargeted Setting} We evaluate the performance of {\m} and the state-of-the-art surrogate-based attacks under the untargeted setting and report the results in Table~\ref{table:mainuntar}. Our results demonstrate that the {\m} outperforms all state-of-the-arts in terms of ASR significantly in just one step under the same perturbation budget. In the following, we elaborate our in-depth observations: (1) Attacks with discriminative surrogates trained on traditionally selected data, namely JDBA and Knock-off demonstrate the worst performance compared to the baselines in all cases. This is because their data selection approaches fail to generate useful and diverse data to capture the target's decision boundaries; 
(2) Attacks with discriminative surrogates trained on the generator data, i.e., DFTA, ST-Data, and DAST outperform the discriminative attacks trained on the traditional data (JDBA and Knock-off). This is owed to their generators, which train more effective data to query the target and train the discriminator. In particular, concentrating on learning the distribution of data close to target's decision boundaries, DFTA and ST-Data illustrate the best performance among the attacks in this category, while DAST demonstrates lower attack success rates due to not restricting the data distribution to data near the decision boundaries; (3) The generative surrogate-based attack, TDB demonstrate the second-best performance and outperforms all the baselines. This indicates that generative surrogates are more effective compared to the discriminative ones as they learn how the joint distribution of inputs and outputs rather than imitating the target's output for a given set of samples; (4) Finally, our proposed {\m} demonstrates the highest ASR compared with all baselines. This is because our attack utilizes a generative surrogate that directly learns the distribution of potential adversarial examples. This distribution contains information about samples residing extremely close/on target's decision boundaries compared to the discriminative surrogates that are trained only on selected samples from this distribution; (5) A notable characteristic of the {\m} is that it achieves the highest ASR in just one attempt (AVG. Steps = 1), which is 200 times less than the average number of steps taken by any other attacks, including the second-best performing baseline. This indicates that our simple attack achieves efficacy and efficiency simultaneously.

\textbf{Targeted Setting} We conduct {\m} and the SOTA attacks in a targeted setting and report the results on CIFAR-10 and CIFAR-100 datasets in Table~\ref{table:maintar}. Our results illustrate that in this more challenging setting, our proposed {\m} outperforms all state-of-the-arts with a notable margin of approximately $15\%$ improvement over the best ASR on average. This is while the {\m} only takes one step to achieve this remarkable success rate, whereas the best-performing baselines require 400 or 500 steps on average, yet is not able to achieve the proposed {\m}'s success rate. The success of {\m} even in this more challenging setting is owed to its generative surrogate that explicitly learns the distribution of samples residing on decision boundaries of any classes, which helps the model find the adversarial example that belongs to any pre-specified class.

%outperform in a more challending setting 
% baselines which consider the decision boundaries achieve better results
% average step of only 1 which is 400 times less 

\subsection{Ablation Study}

Since our attack consists of two steps, we conduct two ablation studies: (1) surrogate ablation study to examine the contributions of %HL: the three criteria of a well-trained 
a generative surrogate on the attack performance; and (2) attack ablation study to examine the effectiveness of our attack by comparing it with different white-box attacks applied on the discriminative substitute network. 
%HL: In the following, we discuss the results of both studies in more detail.

\begin{table}[ht]
\caption{Attack Ablation Study on CIFAR-10 and CIFAR-100 datasets, conducted on Resnet-20 and VGG-19, respectively.} 
\centering
\small
\resizebox{\columnwidth}{!}{\begin{tabular}{l|c|cc|cc}%@{\extracolsep{1pt}}*{3}{c}|@{\extracolsep{1pt}}*{2}{c}}
\toprule   
 && \multicolumn{2}{c}{\textbf{Untargeted}}& \multicolumn{2}{c}{\textbf{Targeted}} \\
\midrule
 & \diagbox{Attack}{Dataset}& CIFAR-10  & CIFAR-100& CIFAR-10  & CIFAR-100\\
 \midrule
  \multirow{5}{*}{\rotatebox{90}{Probability}}&{\m}-P&\textbf{89.8\%(1)}&\textbf{99\%(1)}&\textbf{63.5\%(1)}&\textbf{59.5\%(1)}\\
  \cmidrule{2-2}
  &FGSM-P&46.1\%(1)&51.3\%(1)&26.9\%(1)&17.1\%(1)\\
&PGD-P&73.9\%(261)&82.7\%(257)&50.9\%(345)&44.2\%(309)\\
&C\&W-P&53.1\%(299)&62.2\%(302)&27.1\%(442)&30.4\%(481)\\

\midrule

\multirow{5}{*}{\rotatebox{90}{Label}}&{\m}-L&\textbf{88.8\%(1)}&\textbf{99.2\%(1)}&\textbf{64.3\%(1)}&\textbf{56.5\%(1)}\\
\cmidrule{2-2}
&FGSM-L&49.8\%(1)&52.1\%(1)&39.2\%(1)&31.5\%(1)\\
&PGD-L&75.7\%(262)&85.2\%(264)&51.2\%(351)&44\%(316)\\
&C\&W-L&61.1\%(305)&68.7\%(312)&34.6\%(406)&38.9\%(388)\\

\bottomrule
\end{tabular}}
\label{table:attabl}
\end{table}
  
\textbf{Surrogate Ablation Study} 
We compare the performance of {\m} which uses the generative surrogate with the attack on a discriminative surrogate using the PGD white-box attack. For a fair comparison, the discriminative surrogate is trained on data sampled from our generative surrogate. As shown in Table~\ref{table:ablation}, our proposed attack {\m}, achieves remarkably higher ASRs (approximately +15\% of improvement on average) compared with the attack based on the discriminative surrogate. This demonstrates the 
It is noteworthy that the {\m} achieves this remarkable performance with just one attack iteration whereas the discriminative surrogate requires more than 250 attack iterations.
To further analyze the proposed generative surrogate's behavior, we report the attack performance while adding the three surrogate training criteria (Sec.~\ref{generativesurrogatetraining}) gradually to the ``Base'' surrogate (generative surrogate trained using Eq.~\eqref{eq:gan} and Eq.~\eqref{eq:control} on benign samples). Our results demonstrate that the three criteria based on ``realness'', ``inter-class similarity'', and ``intra-class diversity'' each contribute to the improved performance by forcing potential adversarial examples to look real while possessing the adversarial examples characteristics ($+\mathcal{L}_G^{adv}$), restricting the distribution to samples on/close to the target decision boundaries $+\mathcal{L}^y_{sim}$, while diversifying the samples generated by the surrogate ($+\mathcal{L}^y_{div}$), respectively. The contribution of the intra-class diversity loss ($+\mathcal{L}^y_{div}$) is more evident in the targeted attack, which forces the distribution to spread across as many decision boundaries as possible.

\textbf{Attack Ablation Study} We compare the performance of our proposed attack with different white-box attacks used on the discriminative surrogate (the substitute network in our framework). For the white-box attacks, we use the state-of-the-art FGSM~\cite{goodfellow2014explaining}, PGD~\cite{madry2017towards}, and C\&W~\cite{carlini2017towards} and report the results in Table~\ref{table:attabl}. As the results indicate, our attack outperforms all the white-box attacks applied on the discriminative surrogate, where the PGD attack used as the default attack in all our experiments outperforms the rest. Moreover, in terms of efficiency, our attack takes only one step to achieve the highest success rates. This is approximately 250 times smaller than the multi-step baselines on average. On the other hand, compared with the FGSM, the only white-box attack that is conducted in one step, our attack achieves $40\%$ of improvements on ASR.

\begin{figure}
    \centering
    \subfigure[VGG-16 in CIFAR-10]{\includegraphics[width=9pc, height = 9pc]{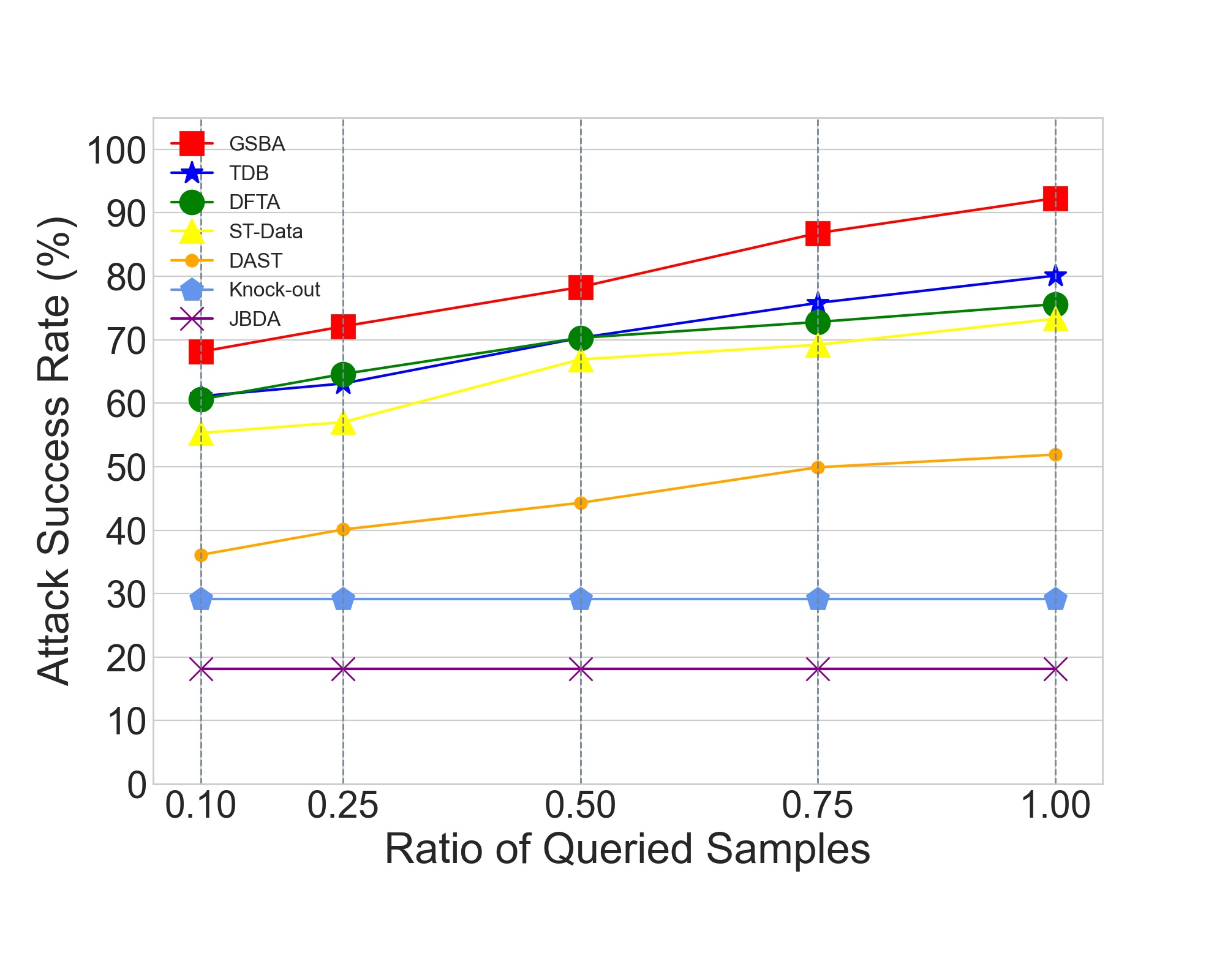}} 
    \subfigure[Resnet-50 in CIFAR-100]{\includegraphics[width=9pc, height = 9pc]{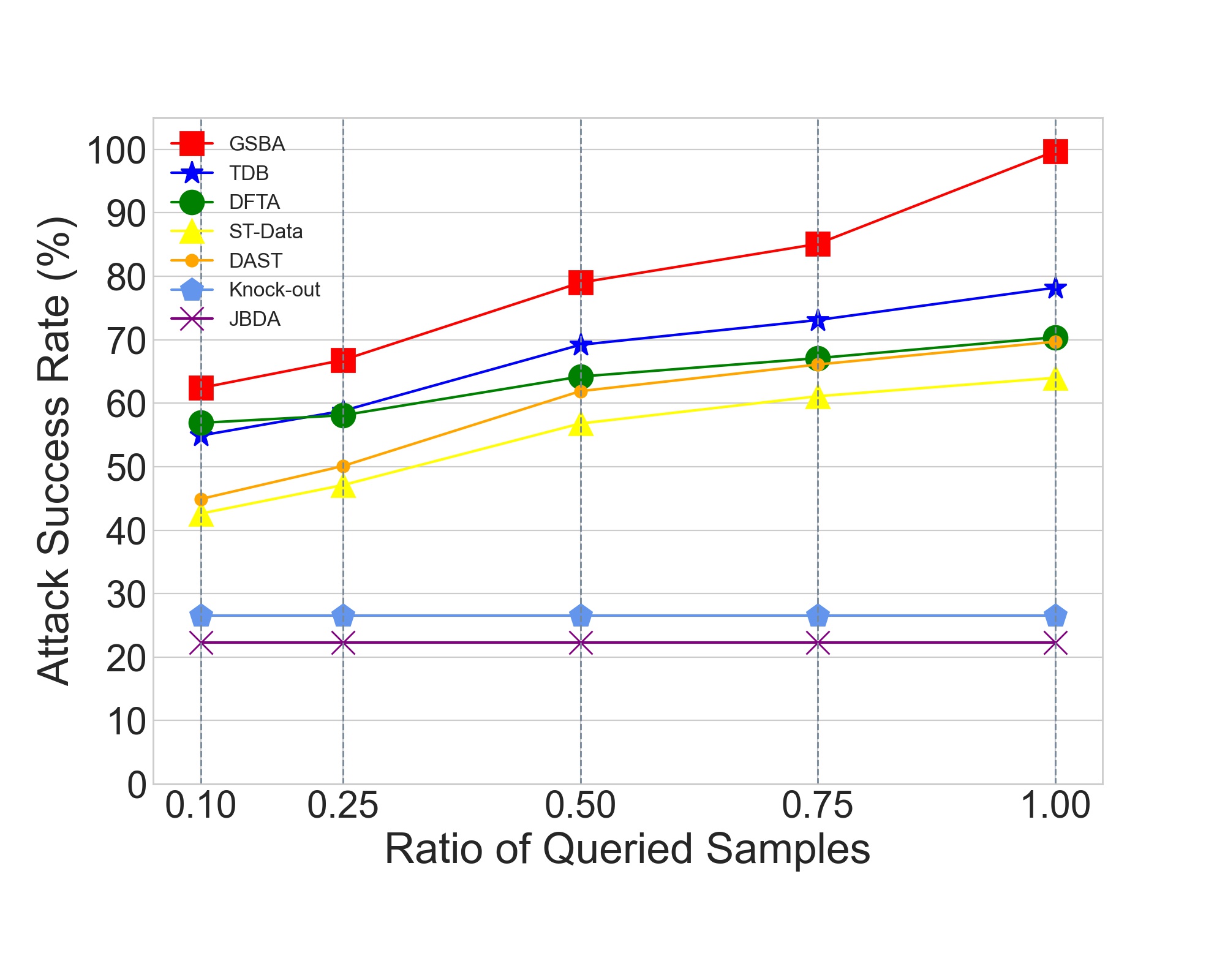}} 
   \vspace{-4mm}    \caption{Attack success rates (ASR) with different query budgets} %HL during the training.}
    \label{fig:querybidget}
\end{figure}

\subsection{Evaluation under Different Query-budgets}
We examine the performance of {\m} under different query-budgets during the training, i.e., the number of samples used to query the target to train the generator or the discriminative surrogates. The results in Figure~\ref{fig:querybidget} are under the query budget of 1 M for {\m} and all baselines. 
%HL: (can we comment this out?) Note that due to the poor performances of Knock-off and JBDA, we only report their ASR with the full query budget (2 M), for the sake of comparison. As shown in the figures, given different query-budgets  {\m} constantly outperforms the baselines. 
Notably, with 50\% of the full query budget, it outperforms the generative-based surrogate attack (TDB) trained on the entire data, which is consistent with the fact that TDB is query intensive during the training since it is designed to learn the entire distribution of the target. Moreover, with 25\% of queried samples, {\m} can achieve comparable performance with the state-of-the-art discriminative surrogate trained with generated data, further confirming that the generative surrogate (generator) is more informative about the adversarial examples. %HL: (can we comment it out) Most remarkably, with only 10\% of the query budget, our proposed {\m} is not only able to outperform the traditional discriminative surrogate-based attacks trained with selected data (JDBA and Knock-off) but achieve approximately 35\% higher success rates compare to the best-performing one. %among 

\iffalse
\subsection{Visualization of Adversarial Examples}
We visualize the adversarial examples generated by {\m}, as well as their corresponding original samples in Figure~\ref{fig:example}. As seen from the Figures, our proposed {\m} can generate adversarial examples with \textit{imperceptible perturbations} by exploiting the distribution of samples residing extremely close to/on target decision boundaries, in just one attempt. This verifies our hypothesis that the generator's distribution is indeed a more informative surrogate for the surrogate-based attacks
\fi
%\subsubsection{Samples of Generated Images}
%\raha{figure}

\section{Conclusion}
 We propose a novel surrogate-based attack that replaces traditional discriminative surrogates with a generative one. Our proposed method directly leverages the distribution learned by this generative surrogate to craft adversarial examples, in just one attempt.
In particular, since the surrogate learns the distribution of samples near or on the target's decision boundaries, which also include adversarial examples, the closest sample to $x_{orig}$ from a different class, drawn from the same distribution, is highly likely to be adversarial. Through extensive experimentation on various benchmarks, our attack demonstrates remarkably higher attack success rates in just one step. Particularly, our attack achieves more than +15\% of improvements over the success rates under the same experimental setup (perturbation budget) as the existing methods. %HL:Our attack is also extremely efficient, as it generates highly successful adversarial examples in just one step.

\bibliographystyle{named}
\bibliography{ijcai24}

\end{document}

\maketitle

\begin{figure}
    \centering
    \subfigure[AlexNet in CIFAR-10]{\includegraphics[width=9pc, height = 9pc]{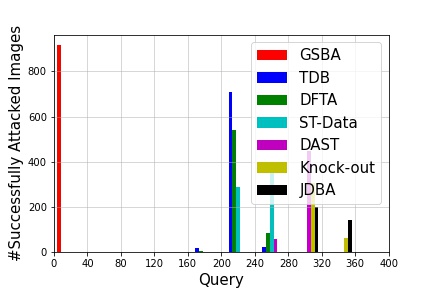}} 
    \subfigure[Resnet-20 in CIFAR-10]{\includegraphics[width=9pc, height = 9pc]{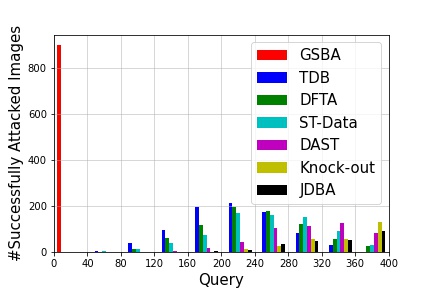}} \\
    \subfigure[Resnet-50 in CIFAR-100]{\includegraphics[width=9pc, height = 9pc]{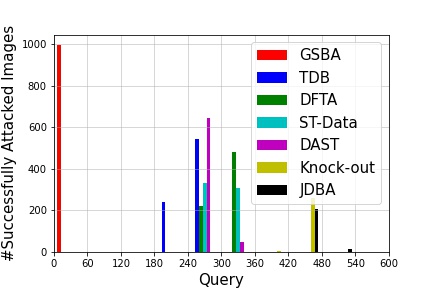}} 
    \subfigure[VGG-19 in CIFAR-100]{\includegraphics[width=9pc, height = 9pc]{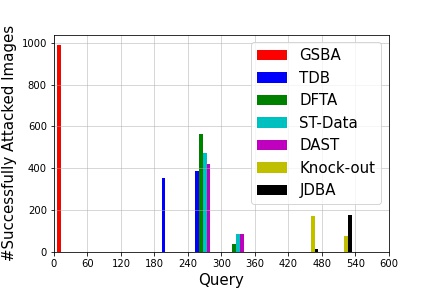}} 
    \caption{ The histogram of step numbers of successful adversarial examples on CIFAR-10 and CIFAR-100 datasets.}%Comparison of the attack success rate at different limited maximum steps in untargeted attack.}
    \label{fig:histquery}
\end{figure}
\section{Implementation Details}

\subsection{Model Architecture and White-box Attack Methods} 
Our framework consists of a generator, a discriminator, and a substitute network. For the generator and discriminator, we use the same architecture used by ~\cite{zhou2020dast}. For the substitute network architecture, we use VGG-13~\cite{simonyan2014very} for the CIFAR-10 and Resnet-18~\cite{he2016deep} for CIFAR-100. It is worth mentioning that our networks do not leverage the pre-trained weights and are trained from scratch. This ensures that no prior knowledge of the target is used to conduct the attack.
For all baselines, we use PGD as our default attack method, unless otherwise mentioned.

Our framework is implemented in PyTorch~\cite{paszke2019pytorch}.  We use ADAM optimizer to train all of our networks. We use mini-batch size of 500 for CIFAR-10 and a bigger mini-batch size of 1000 for the CIFAR-100 dataset. This is because CIFAR-100 dataset has more class categories and thus requires a higher diversity of generated samples. The training hyperparameters in Eq.(3.7), $\alpha_1=1$, $\alpha_2=1$, and $\alpha_3=1$. 
For all baselines, we strictly follow their default experimental setups. 

\subsection{Evaluation Details:} 
We perform our experiments for both targeted and untargeted settings under $l_\infty$ norm. The perturbation bound for each dataset is set to the common choice of $\frac{8}{255}$~\cite{goodfellow2014explaining, ma2021simulating}. 
In the targeted setting, we select the adversarial target class as $y_{adv}=(y_{orig} + 1) ~ \text{mode} ~C$, where $y_{adv}$ is the adversarial target class, $y_{orig}$ is the original class and ``C'' is the total number of classes in the dataset.
Note that we limit the query budget during the training for all methods to 1 M for all methods unless stated differently. For all baselines, we use PGD as our default attack method, unless otherwise mentioned.

\section{Additional Experimental Results}

\subsection{Efficiency of the {\m}}

To analyze the efficiency of the {\m}, we  examine the distribution of steps required for successfully attacked samples. We collect the number of steps used to generate adversarial examples and plot their histogram. We divide the step range into 10 intervals, and count the number of samples in each interval. Our histograms on CIFAR-10 and CIFAR-100 datasets are shown in Figure~\ref{fig:histquery}. Each bar represents one attack and its height indicates the number of samples belonging to the interval (the higher the better). Our results further demonstrate that while existing state-of-the-arts require more than 200 steps to generate successful examples, our attack achieves higher level of success with only one step.

\begin{figure}
    \centering
    \iffalse
    \subfigure[Original on Resnet-20]{\includegraphics[width=9pc, height = 9pc]{Figures/example/gsba-p/grid_image_original_914.jpg}} 
    \subfigure[Adversaril on Resnet-20]{\includegraphics[width=9pc, height = 9pc] {Figures/example/gsba-p/grid_image_adv_p_914.jpg}} \\
   \subfigure[Original on VGG-16]{\includegraphics[width=9pc, height = 9pc]{Figures/example/gsba-p/grid_image_original_537.jpg}} 
    \subfigure[Adversarial on VGG-16]{\includegraphics[width=9pc, height = 9pc] {Figures/example/gsba-p/grid_image_adv_p_537.jpg}} 
  \\
  \fi
    \subfigure[Original on Resnet-20]{\includegraphics[width=9pc, height = 9pc]{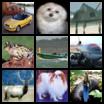}} 
    \subfigure[Adversaril on Resnet-20]{\includegraphics[width=9pc, height = 9pc] {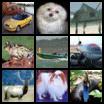}} \\
   \subfigure[Original on VGG-16]{\includegraphics[width=9pc, height = 9pc]{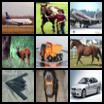}} 
    \subfigure[Adversarial on VGG-16]{\includegraphics[width=9pc, height = 9pc] {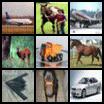}} 
    %\caption{Visualization of adversarial examples generated by {\m}-P and {\m}-L on CIFAR-10. The top two rows represent the examples generated by {\m}-P, and the two last row are the examples by {\m}-L.}
    \caption{Visualization of adversarial examples generated by {\m} on CIFAR-10.}
    \label{fig:example}
\end{figure}

\subsection{Visualization of Adversarial Examples}
We visualize the adversarial examples generated by {\m}, as well as their corresponding original samples in Figure~\ref{fig:example}. As can be seen from the Figures, our proposed {\m} is able to generate adversarial examples with \textit{imperceptible perturbations} by exploiting the distribution of samples residing extremely close to/on target decision boundaries, in just one attempt. This verifies our hypothesis that the generator's distribution is indeed a more informative surrogate for the surrogate-based attacks

%\subsubsection{Samples of Generated Images}
%\raha{figure}

\bibliographystyle{named}
\bibliography{ijcai24}